\documentclass[10pt,twocolumn,letterpaper]{article}

\usepackage{cvpr}              

\usepackage{graphicx}
\usepackage{amsmath}
\usepackage{amssymb}
\usepackage{booktabs}

\usepackage{array}
\usepackage{makecell}

\usepackage{algorithm}
\usepackage{algorithmic}
\usepackage{tabulary}
\usepackage{stfloats}
\usepackage{amssymb}
\usepackage[table,xcdraw]{xcolor}
\usepackage{multirow}
\usepackage{color, colortbl}
\usepackage{wrapfig}

\usepackage{soul}

\usepackage{multirow}
\usepackage[rightcaption]{sidecap}
\usepackage[normalem]{ulem}

\usepackage{times}

\usepackage[pagebackref,breaklinks,colorlinks]{hyperref}

\def\cvprPaperID{768} 
\def\confName{CVPR}
\def\confYear{2022}

\begin{document}

\title{Exploring Geometric Consistency for Monocular 3D Object Detection}

\author{Qing Lian \textsuperscript{1},~~~ Botao Ye\textsuperscript{2,3},~~~Ruijia Xu\textsuperscript{1},~~~Weilong Yao\textsuperscript{3},~~~Tong Zhang\textsuperscript{1}\\
\textsuperscript{1}The Hong Kong University of Science and Technology, \\
\textsuperscript{2}Institute of Computing Technology, Chinese Academy of Sciences, China ~~~\textsuperscript{3} Autowise.AI \\
{\tt\small qlianab@connect.ust.hk, botao.ye@vipl.ict.ac.cn, rxuaq@connect.ust.hk,}\\{\tt\small{ yaoweilong@autowise.ai,  tongzhang@ust.hk }}
}
\maketitle

\begin{abstract}

This paper investigates the geometric consistency for monocular 3D object detection, which suffers from the ill-posed depth estimation. We first conduct a thorough analysis to reveal how existing methods fail to consistently localize objects when different geometric shifts occur. In particular, we design a series of geometric manipulations to diagnose existing detectors and then illustrate their vulnerability to consistently associate the depth with object apparent sizes and positions. 
To alleviate this issue, we propose four geometry-aware data augmentation approaches to enhance the geometric consistency of the detectors. We first modify some commonly used data augmentation methods for 2D images so that they can maintain geometric consistency in 3D spaces. We demonstrate such modifications are important. In addition, we propose a 3D-specific image perturbation method that employs the camera movement. During the augmentation process, the camera system with the corresponding image is manipulated, while the geometric visual cues for depth recovery are preserved.
We show that by using the geometric consistency constraints, the proposed augmentation techniques lead to improvements on the KITTI and nuScenes monocular 3D detection benchmarks with state-of-the-art results. In addition, we demonstrate that the augmentation methods are well suited for semi-supervised training and cross-dataset generalization.


\end{abstract}


\section{Introduction}

\begin{figure}
    \centering
    \subfloat[Visualization of different copy-paste manipulation techniques.]{
    \includegraphics[width=0.85\columnwidth]{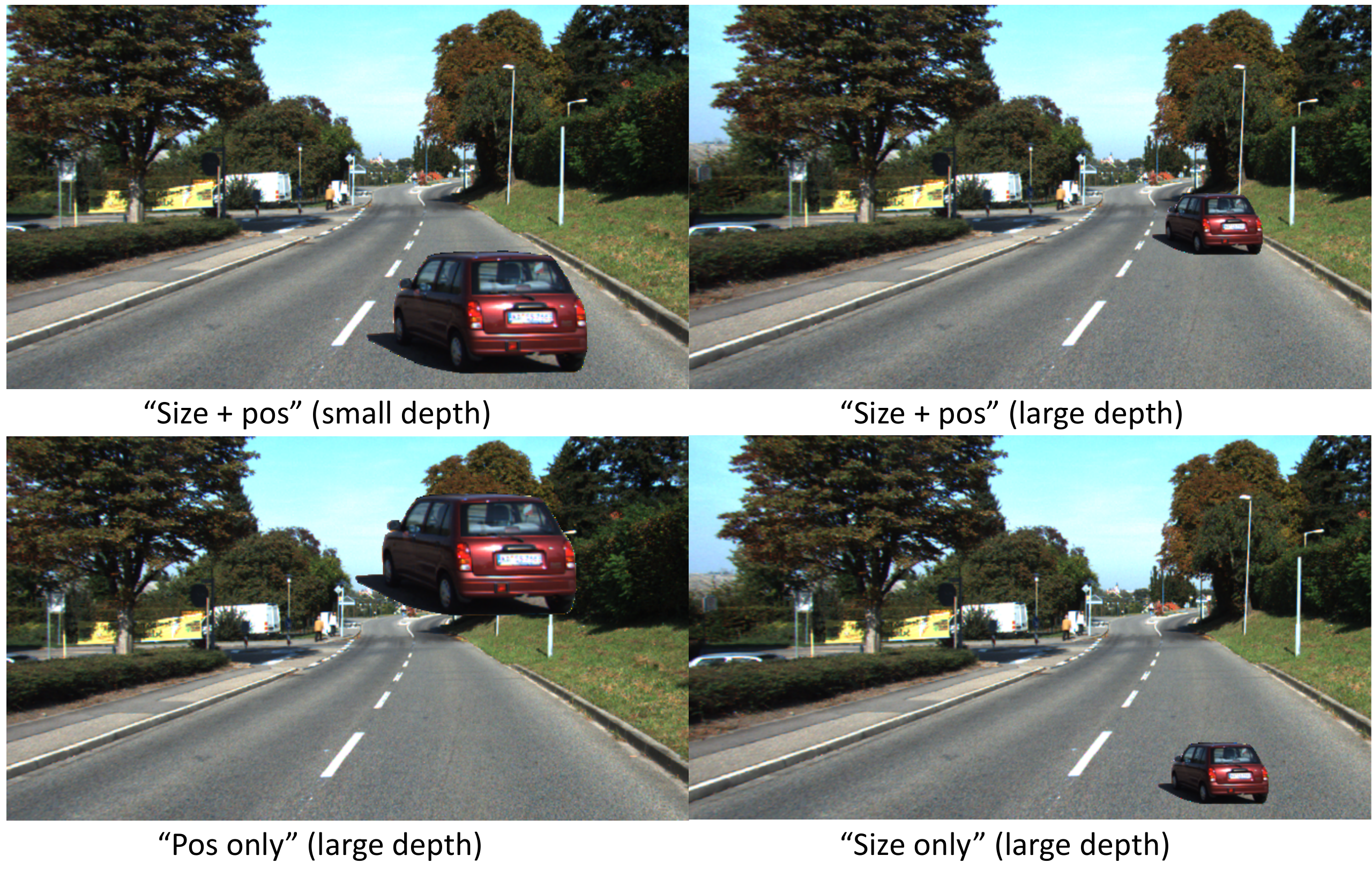}
    \label{fig:fig1a}
    }
    \quad
    \subfloat[Visualization of the estimated depth from the baseline and augmentation-enhanced detectors under the copy-paste manipulation (see the details in Sec~\ref{sec:copy-paste}).]{%
    \label{fig:fig1b}%
    \includegraphics[width=0.9\columnwidth]{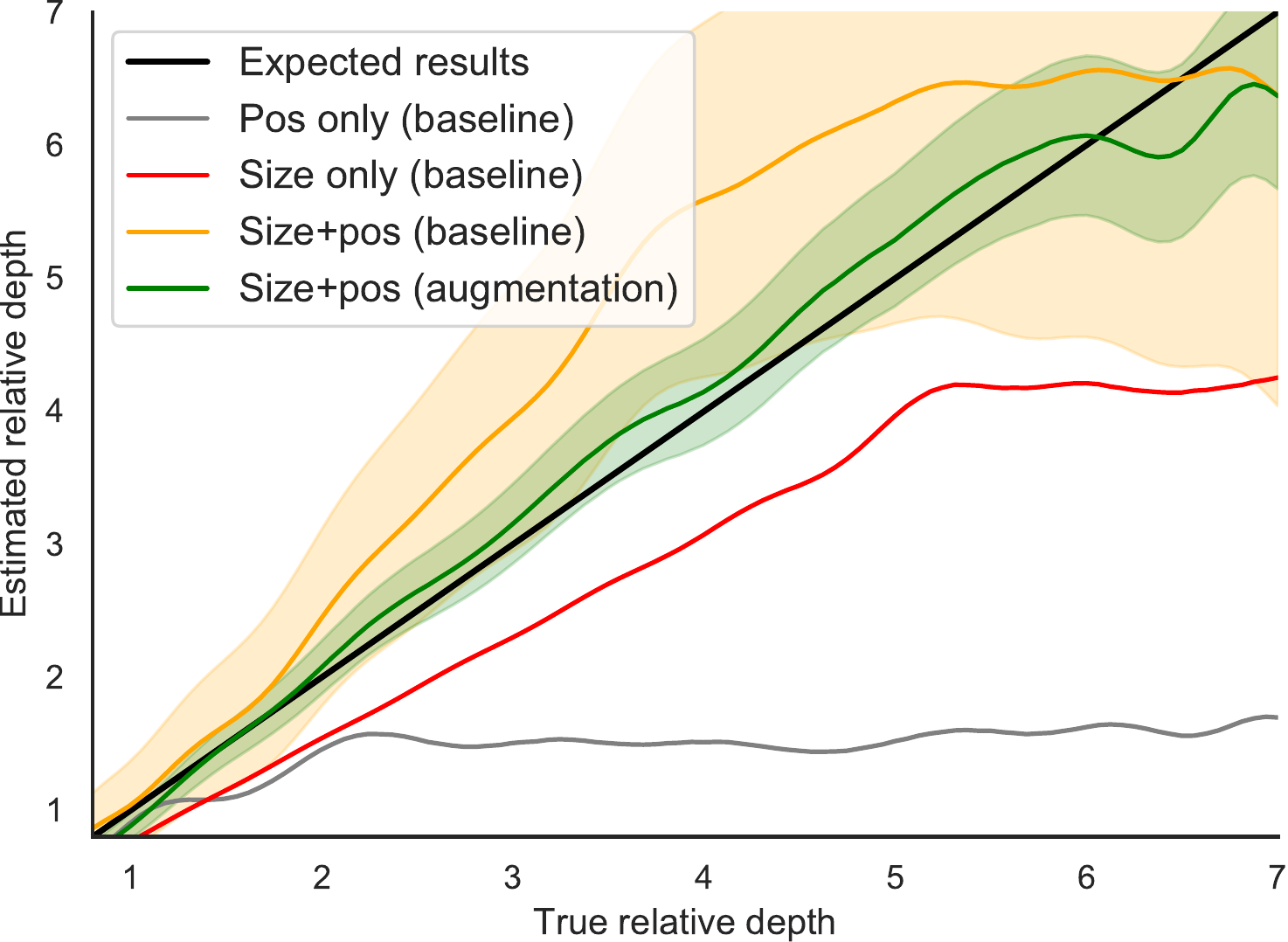}
    }%
    \vspace{-2mm}
    \caption{We select one of the proposed manipulation techniques (copy-paste) to illustrate the instability of object localization under distortion of objects' apparent size and vertical position. ``Size+pos'' denotes geometry-consistent manipulation that shifts the two visual cues with satisfying geometric constraints, ``Size only'' and ``Pos only'' denote geometry-inconsistent manipulation that only shifts the vertical position or apparent size. The shaded region indicates the std of the depth in the ``Size + pos'' manipulation.}
    \label{fig:fig1_all}
    \vspace{-7mm}
\end{figure}
Given an input image, the objective of monocular 3D object detection is to detect objects of interest and recover their position in 3D space.
Recently, it has received increasing attention due to its importance in many downstream tasks, such as autonomous driving, robot navigation, \textit{etc.} 
Different from stereo or lidar sensors, a monocular camera requires a lower cost to perceive the surrounding environments.
However, it suffers from unreliable depth recovery, leading to unsatisfied performance for deployment.

To alleviate the ambiguity in depth estimation, recent approaches~\cite{chen2016monocular, li2019stereo, RTM3D, brazil2019m3d} leverage deep neural networks to model the semantic and geometric information for depth reasoning. However, what geometric features existing detectors use and if they are robust when the used features are perturbed are still under-explored. As a result, this work conducts a comprehensive study on the geometry robustness of existing detectors and proposes several augmentation techniques to enhance their geometric consistency under geometric shifts. 
Different from 2D detection, the geometric visual cues for depth recovery are supposed to be preserved when the objects' coordinates are manipulated, which is not straightforward.

\begin{figure}
    \centering
    \vspace{-3mm}
    \includegraphics[width=0.45\textwidth]{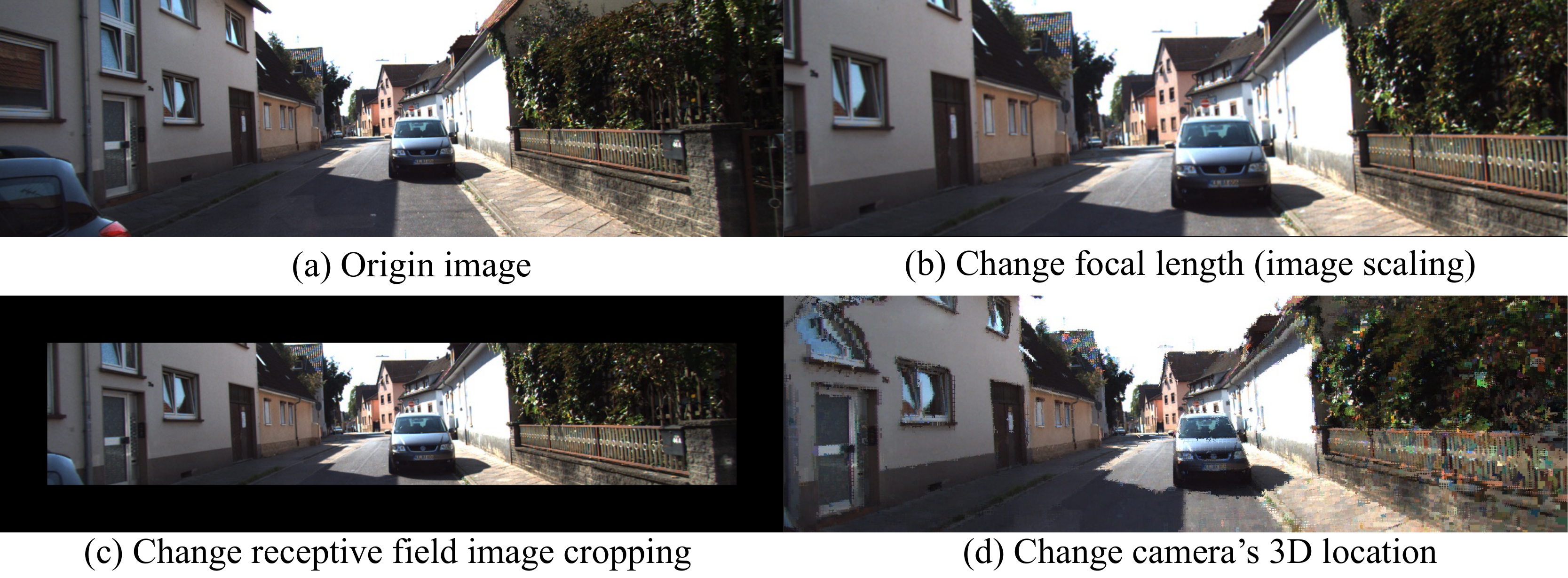}
    \caption{Visualization of the image-level manipulation.}
    \vspace{-2mm}
    \label{fig:vis_image_manipulation}
\end{figure}

It is demonstrated in~\cite{Dijk_2019_ICCV} that neural networks might rely on the features of appearance size and vertical position to estimate object depth.
As visualized in Figure~\ref{fig:fig1a}, objects farther away from the camera have smaller apparent sizes and their vertical position is closer to the vanishing points.
To study if detectors utilize these two pictorial visual cues in localizing objects, we conduct controlled experiments that shift one of the visual cues during manipulating. 
As the results of ``Size + pos'', ``Size only'' and ``Pos only'' shown in Figure~\ref{fig:fig1_all}, the estimated depth changes as the shift of pictorial visual cues, especially for the objects' apparent size. 
We further evaluate the robustness of the detectors in utilizing them to estimate depth by manually distorting the visual cues (\textit{i.e.,} shifting the objects' apparent size or vertical position) with the proposed manipulations (visualized in Figure~\ref{fig:vis_image_manipulation} and~\ref{fig:fig1a}). Through the evaluation, we observe that detectors cannot capture consistent relationships between depth with the two pictorial visual cues, even they can identify the variation of them. 
As shown in Figure~\ref{fig:fig1b} and~\ref{fig:baseline_analysis}, the estimated depth from the baseline detectors has a strong deviation when the images are manipulated. 

Inspired by the above analysis, we convert the manipulations into several geometry-aware data augmentation techniques to improve the geometric consistency of existing detectors. The awareness means that the pictorial visual cues for estimating object depth are preserved during manipulating. At the image level, we lift random scale and random crop, the commonly used 2D augmentation to 3D space by connecting the image manipulation with the shift of camera focal lengths and receptive field. With the help of a dense depth estimation network, we provide a new 3D augmentation method that models the shifts of the camera's 3D location. At the instance level, we propose a geometry-aware copy-paste that leverage the guidance of geometric hints to guide the pasting procedure. Through modeling the geometric constraints, the objects are pasted to novel scenes, while their pictorial visual cues are still preserved. 

By enhancing the geometric consistency, the proposed augmentation techniques yield significant performance boost in both state-of-the-art anchor-free and anchor-based detectors. Compared with the baseline in Figure~\ref{fig:fig1b}, the estimated depth from the enhanced detectors with the designed geometric augmentation methods has less deviation under manipulation. With regularizing the geometric consistency, the trained detectors also show strong robustness in the cross-domain scenario. Furthermore, the consistency regularization techniques also can be applied in the semi-supervised setting, which boosts the performance by regularizing the output consistency under different levels of manipulations. 
Our contributions are summarized as follows:
\begin{itemize}
    \item Through a study of how monocular detectors estimate depth, we identified an instability problem of depth recovery under the changes of the object's apparent size and position.
    \item  We provide four geometry-aware augmentation techniques at the image-level and instance-level to address this problem. 
    With the proposed augmentation techniques, we achieve state-of-the-art results on both the KITTI and nuScenes monocular 3D object detection benchmarks. 
    \item We extend the geometry augmentation techniques into semi-supervised training and cross-domain evaluation, showing the effectiveness of improving performance by regularizing the geometric consistency. 
\end{itemize}

\section{Related work}
In this section, we present the review on monocular 3D object detection and the data augmentation techniques used in object detection. 

\subsection{Monocular 3D detection}
Current monocular 3D object detectors can be split into two categories: image-based and pseudo-lidar based. 

Image-based approaches estimate the 3D information by lifting 2D detectors~\cite{zhou2019objects, 2015_ren_faster} to the 3D space. 
Traditional approaches~\cite{brazil2019m3d, zhou2019objects, simonelli2019disentangling} infer the 3D bounding boxes by additionally estimating location, dimension, and orientation based on 2D detectors~\cite{2015_ren_faster,zhou2019objects}. M3D-RPN~\cite{brazil2019m3d} redesigns the anchor proposal module to better extract 3D information. MonoDis~\cite{simonelli2019disentangling} and MonoFlex~\cite{MonoFlex} address the multi-task learning by disentangling the loss functions and neural network architectures. Shi et al.~\cite{decomp_shi} and Yan et al.~\cite{GuPNet} decompose the depth into two easier estimated metrics: 2D and 3D height. To alleviate the label noise in object location, multiple approaches~\cite{MonoDLE, M3DSSD, MonoEF, decomp_shi, GuPNet} model the aleatoric uncertainty in both the training and inference stages. 
In addition, several methods take external information~\cite{chen2016monocular, qin2019monogrnet, d4lcn, dd3d} (\textit{e.g.,} semantic segmentation, CAD model, the ground surface) to enrich the contextual information for localization.

Except for directly regressing depth, several approaches design 2D and 3D geometry constraints for object depth recovery. 
RTM3D~\cite{RTM3D}, KM3D-Net~\cite{km3d_net}, and MonoPair~\cite{monopair} propose to use the geometric constraints to recovery depth from the constraints in single instance~\cite{RTM3D, km3d_net} or pairwise instances~\cite{monopair}. Similar to MonoPair~\cite{monopair}, RAR-Net~\cite{liu2020reinforced} proposes a reinforcement learning based post-processing strategy to refine the 3D information. To alleviate the sparse constraints, AutoShape~\cite{autoshape_liu} utilizes CAD models to learn dense keypoints to label the semantic keypoints. MonoRun enriches the sparse keypoint constraint to a self-supervised dense constraint, where a modified PnP algorithm is proposed to solve the designed constraint.

In addition to directly taking the monocular image as input,  pseudo-lidar based approaches~\cite{pseudo_lidar, you2019pseudo, pct, DDMP, CaDDN} adopt a depth estimation network~\cite{fu2018deep} to convert the 2D images into 3D point cloud and then apply a point cloud detector on them. 
Although they achieve superior performance, the input transformation requires an extra depth estimation module during inference, leading to high latency.

\subsection{Data augmentation in object detection}
Data augmentation is an effective technique to boost the performance of object detection~\cite{RetinaNet, 2015_ren_faster, zoph2019learning}. Both
geometry-based (\textit{e.g.,} random scale, random crop, and \textit{etc.}) and color-based (\textit{e.g.,} color distortion) augmentation techniques have been widely adopted in 2D detection models~\cite{RetinaNet, 2015_ren_faster, zoph2019learning, zhou2019objects}. In addition, copy-paste augmentation has also proven to be an effective technique to improve the generalization in detection and segmentation. Dvornik et al and Zuo et al~\cite{Dvornik_2018_ECCV, wang2019data} propose to guide the object pasting by aligning the visual context before and after the augmentation. InstaBoost~\cite{instaboost} proposes a probability heatmap to learn where to paste. In the 3D space, Moca~\cite{zhang2020multimodality} proposes an occlusion-aware copy-paste approach for multi-modality 3D detection. In lidar-based detection, data augmentation is also widely adopted~\cite{Cheng2020Improving3O,lang2019pointpillars, mvdepthnet}. Besides the common schemes used in object detection, there are several special augmentation methods tailored to point cloud data, such as the random erasing in SECOND~\cite{yan2018second}, part-aware data augmentation method in~\cite{choi2020part}.

While these aggressive data augmentation methods have yielded impressive gains for either 2D cases or some specific 3D data representation, however, they are hardly leveraged in current monocular 3D detection frameworks due to the violation of geometric constraints, where horizontal flip and color distortion are the only two methods used in this field for a long time. To this end, we hope to reshape this embarrassing situation by offering more diverse geometry-consistent data augmentation techniques to enhance the baseline monocular 3D detectors.

\begin{figure*}[!htb]
\centering 
\includegraphics[width=1.0\textwidth]{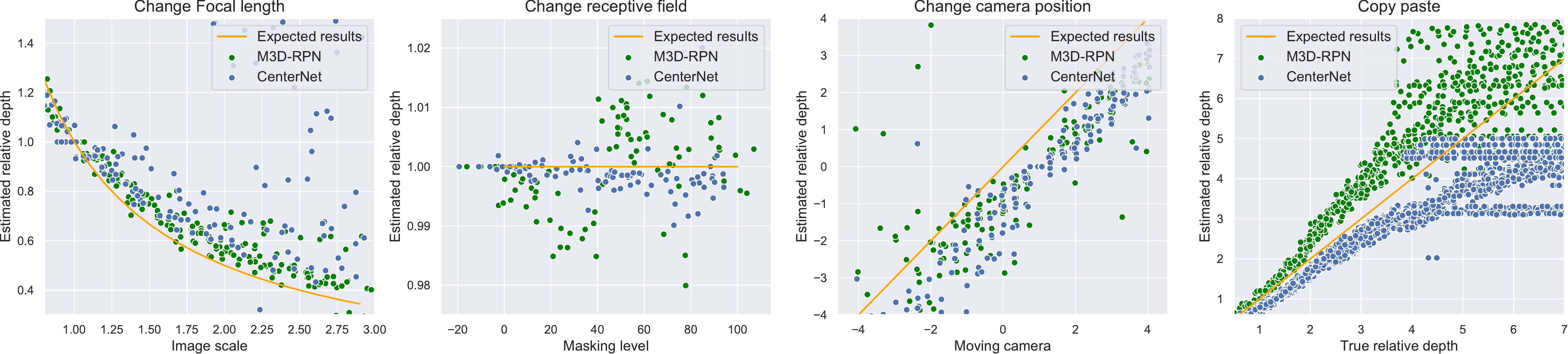}
\caption{Empirical analysis of anchor-based (M3D-RPN) and anchor-free (CenterNet) detectors under geometric manipulations.\label{fig:baseline_analysis} As displayed, their object depth estimation modules are not robust under different geometric manipulations.
}

\end{figure*}

\section{Preliminaries}
\subsection{Baselines}
\label{sec:baseline}

In this section, we first introduce the basic setup of the monocular detectors. We use lower-case and upper-case letters to represent the 2D and 3D coordinates, respectively. Monocular 3D detectors are required to recover the following 3D information: (1) 3D bounding box dimension $[W, H, L]^T$, (2) 3D bounding box center location ${P} = [X, Y, Z, 1]^T$ (3) object yaw angle $\theta$.
On the KITTI dataset~\cite{Geiger2013IJRR}, the following coordinate conversion is adopted to connect the 2D and 3D coordinate:
\begin{align}
    {p} = \frac{1}{Z}K {P},
\end{align}
where ${p} = [u, v, 1]^T$ is the 2D location of the 3D center projected in the image and the transformation matrix $K$ is formulated as: 
\begin{align}
    K = \left(\begin{array}{cccc}
     f & 0 & c_u & 0  \\
     0 & f & c_v & 0 \\
     0 & 0 &1 & 0 \\ 
\end{array}\right).
\end{align}

In this work, we adopt one anchor-free (CenterNet~\cite{zhou2019objects}) and one anchor-based (M3D-RPN~\cite{brazil2019m3d}) detectors as our baselines and lift them to state-of-the-art results by several recently proposed techniques. (1) For depth estimation, we follow~\cite{monopair, MonoDLE, MonoFlex} and model the regression uncertainty with laplacian distribution during training and inference. (2) We add an integral corner loss as in~\cite{MonoFlex, simonelli2019disentangling} to directly supervise the estimated bounding box coordinates with ground-truth. (3) Following~\cite{MonoFlex, smoke, RTM3D, MonoDLE}, we replace the objective of the classification heatmap in CenterNet from the 2D bounding box center to the projected 3D bounding box center. 
\begin{figure}[!htb]
    \centering
    \vspace{-3mm}
    \includegraphics[width=0.45\textwidth]{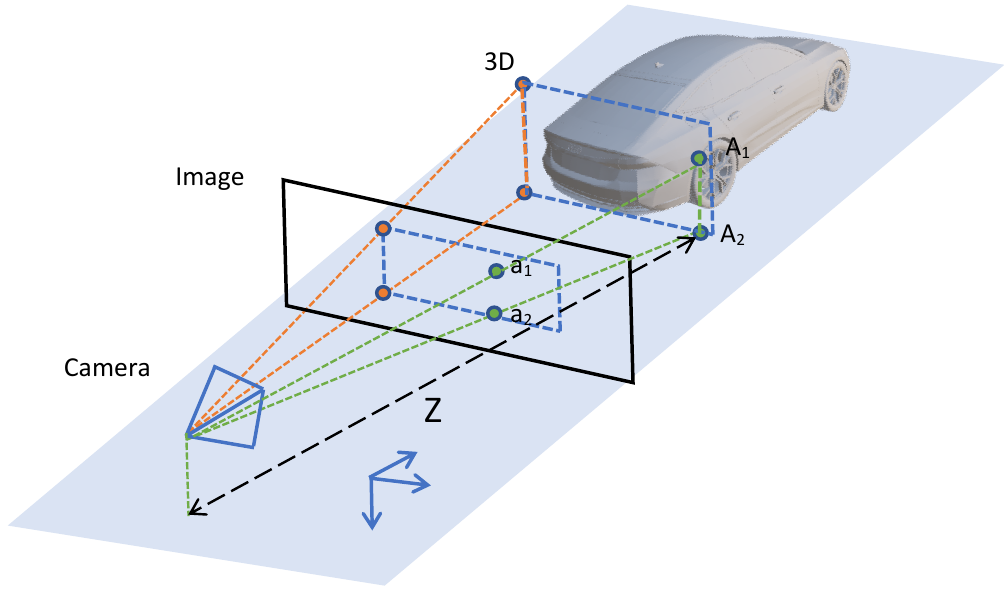}
    \caption{Visualization of the geometric relationships between depth with objects' apparent size and position.}
    \label{fig:vis_relation}
\end{figure}

\subsection{Pictorial visual cues}

In human and machine perception, researchers~\cite{Dijk_2019_ICCV, perception_human} provide several pictorial visual cues that might be used for 3D reconstruction, including object apparent size, vertical position, occlusion, shading, and \textit{etc.} As part of the objective in 3D object detection, the object's apparent size and vertical position are the two most relevant cues for object depth recovery. We visualize the relationships between them with depth in Figure~\ref{fig:vis_relation}. As shown in Figure~\ref{fig:vis_relation}, the orange triangle displays the relationship between 2D bounding box height $h$ and 3D bounding box height $H$ with depth $Z$. Given the camera focal length $f$, we can infer the depth with the following equation:
\begin{align}
    \label{eq:scale_dep}
    Z = f\frac{H}{h}.
\end{align}
The intuition behind this visual cue is that objects that are farther away from the camera tend to have smaller apparent sizes. 

Except for the apparent size, 
depth also can be recovered by localizing the vertical position of the object's ground contact points.
Given the camera height $Y_{cam}$ relative to the ground and the height of the horizon line $v_h$ in the image, depth can be obtained by:
\begin{align}
    Z = f\frac{Y_{cam}}{v - v_h}.
    \label{eq:pos_dep}
\end{align}
In Figure~\ref{fig:vis_relation}, we visualize the relationship of vertical position with depth in the green triangle, where point $A_1$ represents one of the horizon lines projected in the object, point $A_2$ represents one of the object's ground contact points. The points that $A_1$ and $A_2$ projected in image coordinate are $a_1$ and $a_2$, whose vertical positions are $v$ and $v_h$, respectively. The intuition behind this visual cue is that an object closer to the camera would have a lower vertical position in the image. Although the two geometric relationships require several assumptions, most of them are satisfied in autonomous driving environments. We refer readers to~\cite{Dijk_2019_ICCV} for a more thorough review of the pictorial cues.


\section{Analysis based on Geometric manipulations}

In this section, we first present three image-level and one instance-level geometric manipulation techniques to disturb the aforementioned visual cues in the image. Then we introduce the robustness analysis based on the presented manipulation techniques. KITTI validation set~\cite{chen20153d} is adopted to conduct the empirical analysis.

\subsection{Image-level}

\noindent{\textbf{Random Scale.}}
\label{sec:focal_length}
Random scale resizes the image with a specific scale, which corresponds to shifting the camera focal length in the imaging process. Under the same camera intrinsic in the pinhole camera, image scaling also can be treated as moving all the objects towards a relative scale. For a scaling factor $s$, the location change in 3D space is  formulated as:
\begin{align}
    {P_{new}} = \left(\begin{array}{cccc}
     1 & 0 & (1-s)\frac{c_u}{f} & 0 \\ 
     0 & 1 & (1-s)\frac{c_v}{f} & 0 \\
     0 & 0 & s & 0 \\
     0 & 0 & 0 & 1 
\end{array}\right){P}.
\end{align}
We evaluate if the detector can identify this location change when the objects are scaled with different sizes. 

\noindent{\textbf{Random Crop.}}
\label{sec:cam_field}
The second manipulation is randomly cropping the image, which corresponds to changing the camera receptive field. To preserve the pictorial visual cue during manipulating, we pad the cropped region to keep the objects' vertical position in random scale. As demonstrated by Md et al~\cite{Md2020Position}, neural networks would utilize the padding region to extract the position information. We evaluate if the detectors are robust under this manipulation technique by checking if they can estimate consistent depth after cropping and padding.

\begin{table}[!htb]
    \centering
    \caption{Experimental results of anchor-based (M3D-RPN) and anchor-free (CenterNet) detectors under different manipulation techniques. Except the baseline, we replace the ground-truth with estimated results. For example, ``Depth*`` denotes replacing the ground truth depth with the estimation and setting all other components with ground truth. (Results of $AP|_{40}$ with IoU$\geq$0.5 on car (easy) are reported.)}
    \tabcolsep2pt
    \begin{tabular}{ll|ccccc} \hline
    Network&    Method & Base & Depth* & Dim* & Pos*  \\ \hline
    \multicolumn{1}{c}{\multirow{5}{*}{M3D-RPN}} &  
    Origin &  54.3 & 55.6 & 99.1 & 98.9 &\\ 
        &Random scale &31.3 & 34.8 & 98.2 & 98.4    \\ 
        &Random crop & 40.2 & 42.3 & 95.6 & 96.7\\
        &Moving cam & 25.6 & 29.4 & 91.0 & 89.3\\
        & Copy-paste & 35.2 & 43.3 & 83.4 & 97.3\\ \hline
    \multicolumn{1}{c}{\multirow{5}{*}{CenterNet}} & 
        Origin  & 49.9 & 50.6 & 98.9 & 99.0 \\
        & Random scale & 23.3 & 27.3 & 97.8 & 97.9  \\ 
        & Random crop& 38.8 & 41.0 & 94.7 & 94.2 \\
        & Moving cam & 25.9 & 28.8 & 91.7 & 88.6 \\
        & Copy-paste & 36.2 & 42.3  & 82.0 & 97.0 \\ \hline
    \end{tabular}
    \vspace{-5mm}
    \label{tab:empirical_analysis}
\end{table}

\noindent{\textbf{Moving Camera}}
\label{sec:cam_pos}
The third manipulation is moving the camera's location, which equals to taking images from a different location.
In this manipulation, we change the camera's location in the $Z$ coordinate, where the object-to-camera distance should be shifted with an offset $d$:
\begin{align}
    {P_{new}} = {P} + [0,0,d,0]^T.
\end{align}
To generate corresponding images, we adopt a depth estimation network: DORN~\cite{fu2018deep} to regress the location of each pixel. With the manipulated images, we evaluate if the detectors cannot identify the offset in the generated image.  

\subsection{Instance-level: Copy-paste}
\label{sec:copy-paste}
In addition to the image-level manipulation, we further provide an instance-level manipulation: copy-paste. Copy-paste is widely used in 2D instance segmentation, where several approaches are proposed to preserve the semantic context during pasting. However, most of the approaches~\cite{Dvornik_2018_ECCV, instaboost, wang2019data} ignore the geometric relationships, destroying the pictorial visual cues during manipulation. 
We first provide a geometric consistent copy-paste to study the robustness of the detectors and then introduce two geometric violated copy-paste to study how neural networks estimate depth.

\noindent{\textbf{Geometric consistent copy-paste}}
This manipulation is split into two stages: (1) what to copy and (2) how to paste.

\noindent{\textbf{What to Copy.}}
In this stage, we first collect an instance database from the training data. Specifically, we crop the objects of interest in the training images by a pre-trained instance segmentation model~\cite{wu2019detectron2}. To filter out outliers, we remove the instances that are truncated or have low visibility. Since the two pictorial visual cues we studied assume the ground is flat,  we further remove the unqualified objects by comparing their corresponding vanish points as in~\cite{Dijk_2019_ICCV}.

\noindent{\textbf{How to Paste.}}
In the pasting stage, we sample depth in a valid region (\textit{i.e.,} [0m, 60m]) and then calculate the corresponding bounding box size and the pasting location based on Equation~\ref{eq:scale_dep} and~\ref{eq:pos_dep}. The whole pipeline of pasting is described in Algorithm~\ref{alg1}.


\begin{algorithm}
  \begin{algorithmic}[1]
    \STATE \text{\textbf{Input}:Original object with ground truth:}\\ 
    \text{$\quad [(u_1, v_1, u_2, v_2), (X, Y, Z), (W, H, L), \theta]$.}
    \STATE \text{Sample a new scene for pasting.}
    \STATE \text{Sample new depth $\hat{Z}$.}
    \STATE \text{Set the orientation $\hat{\theta} = \theta$.}
    \STATE \text{Set the location of $\hat{X} = X\frac{\hat{Z}}{Z}$.} 
    \STATE \text{Compute the location of $\hat{Y}$ based on Eq~\ref{eq:pos_dep}.}
    \STATE\text{Set the dimension as $\hat{W} = W, \hat{H} = H, \hat{L} = L$.}
    \STATE \text{Generate a 2D bounding box $(\hat{u_1}, \hat{v_1}, \hat{u_2}, \hat{v_2})$ by }\\ \text{ projecting the corner points in 3D boxes to the image.}
    \IF{the new instances does not satisfy the Eq~\ref{eq:scale_dep}.} 
    \STATE Go back to Step 2.
    \ENDIF
    \STATE \text{\textbf{Output}: the new instances with ground truth: } \\ 
    \text{$\quad[(\hat{u_1}, \hat{v_1}, \hat{u_2}, \hat{v_2}), (\hat{X}, \hat{Y}, \hat{Z}), (\hat{W}, \hat{H}, \hat{L}), \hat{\theta}]$.}
  \caption{Procedure of copy-paste augmentation. }
  \label{alg1}
  \end{algorithmic}

\end{algorithm}

Note that to simplify the generation process,  we fix the object yaw and alpha angle during pasting. Step 4 and Step 5 display how we use the geometric relationship to determine the objects' apparent size and vertical position. For the geometry violated manipulation, the value in step 3 and step 5 are randomly sampled. The if statement in step 9 would be false when the height of the ground plan in the origin and pasted scenes are different. Figure~\ref{fig:vis_copy_paste} visualizes the difference between geometry consistent and geometry violated copy-paste.

\subsection{Stability under different manipulations}

In Figure~\ref{fig:baseline_analysis}, we plot the estimated depth of the detectors for manipulated images and compare it with the expected depth to measure whether the detectors are robust against the four above-mentioned manipulations. As illustrated, while the estimated depth in anchor-based and anchor-free detectors is approximately correlated with the expected result, however, both of them suffer from a large deviation, especially for the anchor-free detector.
To further evaluate if the detectors can capture the variation of each visual cue and learn consistent geometric relationships, we report the mAP with the prediction of depth, 3D dimension and position in Table~\ref{tab:empirical_analysis}. As illustrated, the \textit{base} version denotes the overall mAP with the estimation results. The versions of  \textit{depth*}, \textit{dim*} and \textit{pos*} mark the mAP with the estimated depth, dimension and position offset respectively, while leaving the other components the same as the ground truth.  We draw the following observations: 1) In the origin setting, the performance drop in \textit{depth*} is larger than \textit{dim*} and \textit{pos*}, showing that the depth recovery is more challenging; 2) Both detectors suffer from a significant performance drop under the four kinds of manipulations, especially for the anchor-free detector; 3) For the results of \textit{dim*} and \textit{pos*}, they almost achieve 100\% mAP, showing that the detectors accurately estimate the dimensions and positions of the objects, even in the manipulated image. However, the accuracy in \textit{depth*} is much less than 100\%, indicating that the detectors cannot capture consistent geometry relationships under the manipulations; 4) Unlike the phenomenon in the image-level manipulation, detectors are unable to accurately regress the objects' dimensions for the inserted objects.

\section{Geometry-aware data augmentation}
\label{sec:aug}
After diagnosing the geometric inconsistency of the detectors, we convert the manipulations into geometric consistent augmentation approaches to enhance this consistency. 

\noindent{\textbf{Random Scale}}
As aforementioned in Section~\ref{sec:focal_length}, we distort the camera focal length to generate the image with  scales from 0.8 to 1.2. Although images with different scales are generated, cameras' intrinsic may be inconsistent with the testing data, which would be harmful to the testing performance. To customize the detectors with this augmentation method, we disentangle the training objective of depth from $Z$ to a camera intrinsic irrelevant $\frac{Z}{f}$. During inference, we recover the depth by timing $\frac{Z}{f}$ with the corresponding camera focal length. For the other 3D metrics, we fix them as the original value, because they are consistent under different image scales.   

\noindent{\textbf{Random Crop}}
As discussed in Section~\ref{sec:cam_field}, we adopt a crop-then-pad operation to make sure the geometric cue is consistent during training and inference. We randomly cropped out 25\% of the region and adopt a zero-padding to fill the image in the vertical direction. 

\noindent{\textbf{Moving camera}}
Regarding moving the camera, we randomly move the camera in the Z direction with a range from -5m to 5m. For the coordinate conversion in the 2D and 3D coordinates, we adopt the same operation as in Section~\ref{sec:cam_pos}.  To simplify the augmentation process, we do not adopt sophisticated novel view synthesis models, while leveraging neural networks to convert the pixel in the origin view to the target view. For the pixels that cannot find the corresponding pixel in the source view, we fill them by the nearest neighbor pixel.  

\begin{figure}[!htb]
    \centering
    \vspace{-2mm}
    \includegraphics[width=0.4\textwidth, trim=0 190 230 0, clip]{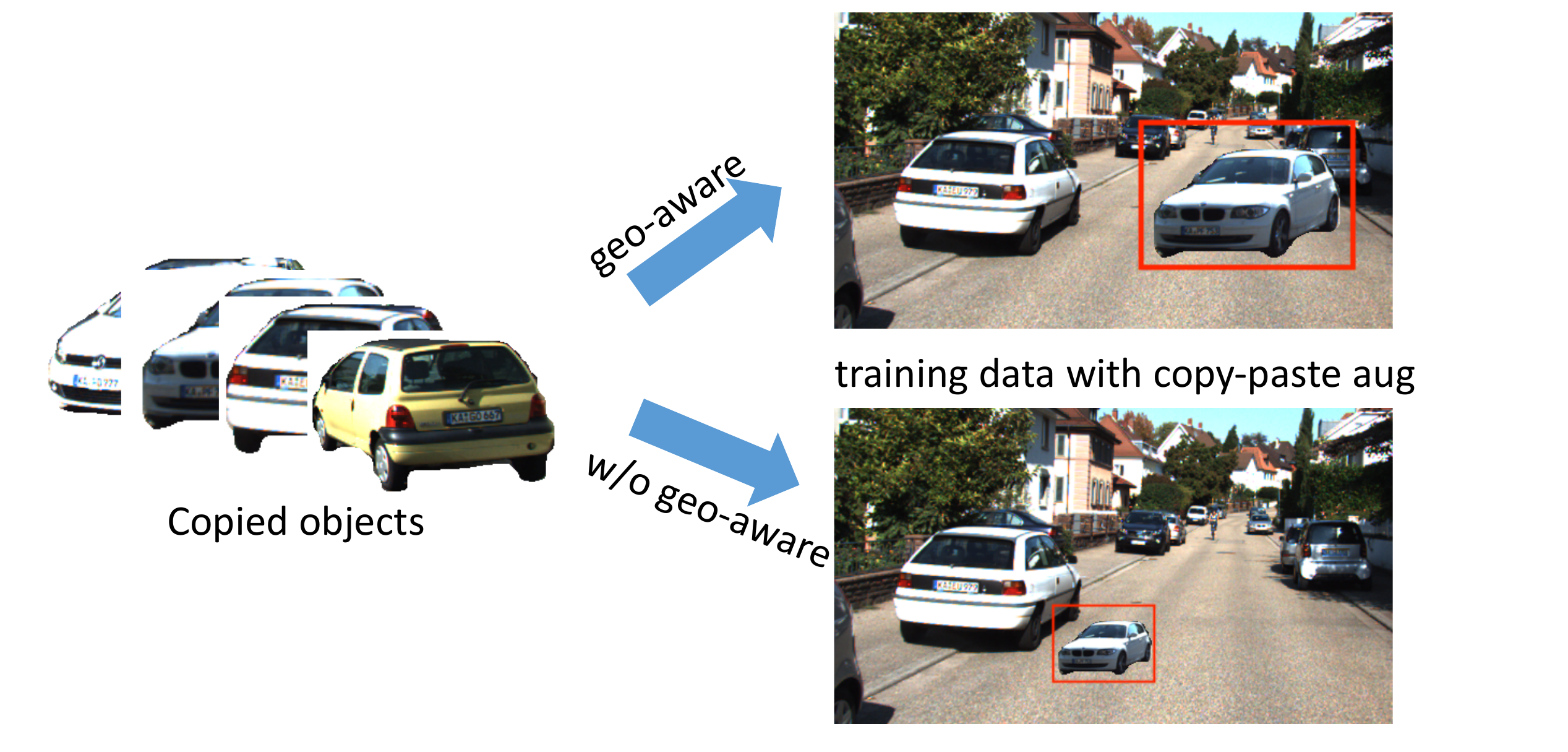}
    \caption{Visualization of copy-paste data augmentation with and without geometry-aware consideration.}
    \vspace{-4mm}
    \label{fig:vis_copy_paste}
\end{figure}

\noindent{\textbf{Copy-Paste}}
For the copy-paste augmentation technique, we adopt the geometric-consistent version as discussed in Section~\ref{sec:copy-paste}. As visualized in Figure~\ref{fig:vis_copy_paste}, the apparent size and vertical position are matched with ground truth depth after considering the geometry relationships.

\label{sec:analysis}

\section{Experiments}

We first introduce the experimental setup, including evaluation benchmarks, metrics, and our implementation details. Then, we present and analyse the results of our experiments. In addition, we verify the effectiveness of the proposed augmentation techniques in label-efficient settings.

\subsection{Experimental setup}
We evaluate the effectiveness of the proposed data augmentation approaches on the KITTI~\cite{geiger2012we} and nuScenes~\cite{nuscene} 3D object detection benchmarks. 

\noindent{\textbf{KITTI}}~\cite{geiger2012we} consists of 7,481 training frames and 7,518 test frames with 80,256 annotated 3D bounding boxes. 
For fair comparisons, we follow prior work~\cite{chen20153d, chen2016monocular} and split the training data into training and validation subsets. 
We evaluate the effectiveness of the proposed components on the validation set and evaluate the final model on the test set.

\noindent{\textbf{nuScenes}}~\cite{nuscene} is a recently released autonomous driving dataset. 
It contains up to 40K annotated key frames from 6 cameras with 4 different scene locations. 
Compared with the KITTI dataset, it has 7x as many annotations with 23 different object classes. The dataset is split into 700 video sequences for training, 150 for validation, and 150 for testing. 
Due to the limited computation resources, we train the detectors on the training subset and evaluate the performance on the official validation subset. 

\noindent{\textbf{Evaluation metrics}}
In the KITTI dataset, we follow the official protocol~\cite{geiger2012we} and adopt the $AP|_{40}$ evaluation metrics on both bird-eye view (BEV) and 3D bounding box estimation tasks. The evaluation is conducted separately based on the difficulty levels (Easy, Moderate, and Hard) and object categories (Car, Pedestrian, and Cyclist). In the nuScenes dataset, we adopt the provided~\cite{nuscene} evaluation metrics from the perspective of entire boxes (mAP), translation (mATE), size (mASE), \textit{etc.}

\noindent{\textbf{Implementation details}}
As described in Section~\ref{sec:baseline}, our experiments are conducted based on  CenterNet~\cite{zhou2019objects} and M3D-RPN~\cite{brazil2019m3d}.  We use the modified DLA-34~\cite{zhou2019objects} (CenterNet) and DesNet-141~\cite{huang2017densely} (M3D-RPN) as detectors' backbone and initialize the parameters with ImageNet~\cite{deng2009imagenet} pre-trained weights. 
Before applying the proposed augmentation techniques, we first pad the images in KITTI to the size of 1280$\times$384 and downsample the images in nuScenes to half of the resolution. Regarding optimization, we train the two detectors with 90 epochs in the KITTI dataset and 12 epochs in the nuScenes dataset. We adopt the AdamW~\cite{loshchilov2018adamw} optimizer for training and set the initial learning rate as 3e-4. The detailed descriptions of the experimental setup are provided in the supplementary material.  

\subsection{Individual and composite effect of the proposed augmentation methods}
To evaluate the effectiveness of our geometry-aware strategy, we first conduct experiments with different augmentation strategies for comparison. In the vanilla strategy, we adopt the horizontal flip augmentation in both 2D and 3D tasks. For the other augmentation techniques (random scale, random crop, and copy-paste), we only adopt them in the 2D task, because the vanilla operations violate the geometric constraints and cannot directly get the corresponding 3D ground-truth.  In our geometry-aware scheme, we add the coordinate-based augmentation to 3D task with the proposed geometric-preserving operations, where the 3D related ground-truth are calculated as in Section~\ref{sec:aug}. Table~\ref{tab:ablation_aug} displays the comparison results with anchor-based (M3D-RPN) and anchor-free (CenterNet) detectors. As illustrated, the geometry-aware scheme consistently improves the vanilla strategy and the combination of four augmentation techniques yields consistently performance boosting with 5.99\%/4.79\%, 4.96\%/3.75\%, and 3.85\%/2.35\% of the three settings on the two detectors, respectively. We also observe that the improvement of ``vanilla aug'' over ``w/o aug'' is limited. The potential reason is that the performance of monocular 3D detection heavily relies on the accuracy of depth recovery,  while vanilla augmentation destroys the pictorial visual cue for recovery. 
\begin{table}[htbp]
  \centering
  \vspace{-2mm}
  \caption{Comparison among different augmentation strategies on the KITTI validation dataset. $AP|_{40}$ of 3d bounding box on the Car category are reported.}
    \begin{tabular}{c|l|ccc} \hline
    \multicolumn{1}{l|}{Method} & Setting & Easy  & Mod   & Hard \\ \hline
    
    \multirow{7}[0]{*}{\makecell{M3D-RPN}} 
          & W/o aug & 17.45 & 10.03 & 9.42 \\ 
          & Vanilla aug & 18.21 & 11.28 & 9.56 \\
          & + Random scale & 22.06  &  15.43     & 12.04 \\
          & + Random crop &  20.91 & 14.42 & 11.60 \\
          & + Moving cam & 21.73 & 14.56 & 11.37 \\ 
          & + Copy-paste &   22.63   &   15.94    & 12.61 \\
          & All aug &  \textbf{23.42} &  \textbf{16.24} & \textbf{13.41}  \\ \hline
    \multirow{7}[0]{*}{\makecell{CenterNet}}
          & W/o aug & 18.74 & 13.21 & 10.80  \\ 
          & Vanilla aug & 20.16 & 13.49 & 11.95 \\
          & + Random scale & 22.46 & 15.60  & 13.57 \\
          & + Random crop & 22.63 & 16.02 & 13.21 \\
          & + Moving cam & 21.34 & 15.10  & 12.92 \\
          & + Copy-paste & 22.23 & 15.47 & 13.24 \\ 
          & All aug & \textbf{24.53} & \textbf{17.23} & \textbf{14.32} \\ \hline
    \end{tabular}%
  \label{tab:ablation_aug}%
\end{table}%

\subsection{Results on the KITTI test set}
In Table~\ref{tab:test_kitti}, we present the comparison of the  proposed augmentation enhanced detectors with state-of-the-art methods on the KITTI test set. Quantitatively, the two baseline approaches with vanilla augmentation already achieve comparable results in each setting. Powered by the proposed geometry-aware augmentation, we outperform the baseline with 3.89\%/4.00\%, 3.03\%/2.05\%, and 2.00\%/1.76\% of the three different difficulties in the 3D task. For the anchor-based detectors, we outperform the state-of-the-art approach DDMP-3D~\cite{DDMP} a large margin while keeping a low running time. For the anchor-free detector, we achieve almost 2\% improvement over the state-of-the-art  method MonoEF~\cite{MonoEF}. 

\begin{table*}[htbp]
  \centering
\caption{Experimental results of the ``Car'' class on the KITTI Test set. The best results are marked with \textbf{bold}.}
    \begin{tabular}{ll|cccccc|c} \hline
  \multicolumn{2}{c|}{\multirow{2}{*}{Setting}}& \multicolumn{3}{c}{3D (Test)} & \multicolumn{3}{c|}{BEV (Test)} & \multirow{2}{*}{Running time (ms)} \\
 &  & Easy & Mod & Hard & Easy & Mod & Hard \\ \hline
 \multirow{8}{*}{Anchor-based}    &  M3DSSD & 17.51 & 11.46 & 8.98  & 24.15 & 15.93 & 12.11 & - \\
  &  Mono R-CNN & 18.36 & 12.65 & 10.03 & 25.48 & 18.11 & 14.10& 70 \\
   & GrooMed-NMS & 18.10  & 12.32 & 9.65  & 26.19 & 18.27 & 14.05 & - \\
    &   Kinemantic & 19.07 & 12.72 & 9.17  & 26.69 & 17.52 & 13.10 & - \\
   &  MonoRun & 19.65 & 12.30  & 10.58 & 27.94 & 17.34 & 15.24 & 70\\
 &   DDMP-3D  & 19.71 & 12.78 & 9.80  & 28.08 & 17.89 & 13.44 & - \\  
   &  CaDDN & 19.17 & 13.41 & 11.46 & 27.94 & 18.01 & 17.19 & 630\\

\cline{2-9}
   &  M3D-RPN (vanilla aug) &  16.45  &  11.24   & 10.02  &  26.53     &     17.78  &   12.11 & 40 \\
    & M3D-RPN (geo aug) & \textbf{20.34}     &   \textbf{14.27}    &  \textbf{12.02}     & \textbf{28.15}      &  \textbf{19.67}     & \textbf{16.73} & 40\\ \hline
   \multirow{6}{*}{Anchor-free}&  MonoFlex & 19.94 & 13.89 & 12.07 & 28.23 & 19.75 & 16.89 & 30 \\
   & MonoEF & 21.29 & 13.87 & 11.71 & 29.03 & 19.70 & 17.26 & 30\\
    & AutoShape & 22.47 & 14.17 & 11.36 & 30.66 & 20.08 & 15.59 & 50\\
   &  Monodle & 17.23 & 12.26 & 10.29 & 27.94 & 17.34 & 15.24 & 40\\
    \cline{2-9}
   &  CenterNet (vanilla aug) & 19.41      &  13.21     &   11.04    &  27.89     &  19.24     &  15.53 & 30\\
  &   CenterNet (geo aug) & \textbf{23.41} & \textbf{15.26} & \textbf{12.80} & \textbf{31.58} & \textbf{20.75} & \textbf{17.66} & 30\\ \hline
    \end{tabular}%
  \label{tab:test_kitti}%
\end{table*}


\begin{table}[!htb]
    \vspace{-3mm}

    \centering
    \caption{Experimental results of the anchor-free detector on the nuScenes validation set. }
\vspace{-3mm}

    \begin{tabular}{l|cccc} \hline
        Setting & mAP$\uparrow$&mATE$\downarrow$&mASE$\downarrow$& NDS$\uparrow$ \\ \hline
        Vanilla aug & 33.2 &  0.69 & 0.28  & 38.4      \\  
        Geo aug & 34.5   & 0.68 & 0.27   & 39.4      \\ \hline
    \end{tabular}
    \label{tab:nuScenes}
        \vspace{-3mm}
\end{table}

\subsection{Results on the nuScenes dataset}
Except for the KITTI dataset, we also evaluate the proposed augmentation techniques on the nuScenes dataset. Table~\ref{tab:nuScenes} presents the experimental results of the modified CenterNet on nuScenes validation set.  Although nuScenes contains more training instances, the proposed geometry-aware augmentation strategies still improve the vanilla setting in different evaluation metrics. Typically, regarding the most important mAP metric, the geometry-aware strategy outperforms the vanilla version over 3.89\%.
\begin{figure}[!htb]
    \centering
    \vspace{-5mm}
    \includegraphics[width=0.4\textwidth,  trim=0 30 30 0,]{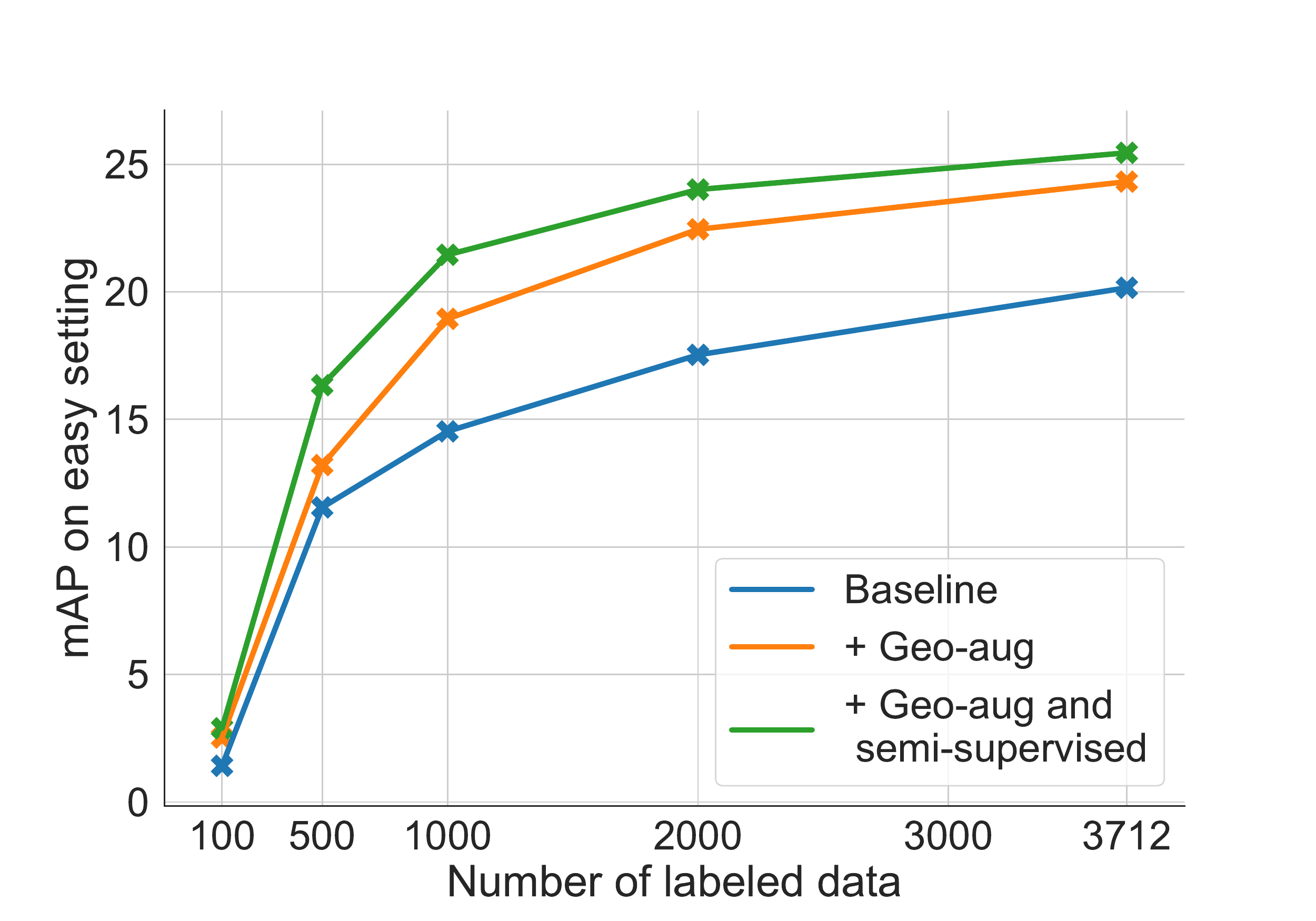}
    \caption{Experimental results of our geometric data augmentation on the semi-supervised learning setting.}
    \vspace{-5mm}
    \label{fig:semi}
\end{figure}

\begin{table}[!htb]
    \centering
    \caption{Cross-domain evaluation between different augmentation methods with the anchor-free detector. Results of car on the KITTI (easy with 3D mAP) and nuScenes datasets (mAP) are reported. }
    \vspace{-1mm}
    \label{tab:cross_domain}
    \begin{tabular}{c|ccc} \hline
     Training data & Setting  & KITTI & nuScenes \\ \hline
     \multicolumn{1}{c|}{\multirow{2}{*}{KITTI}}    &  Vanilla aug & 20.16 & 10.23 \\ 
                                                  & Geo aug & 24.53 & 19.40  \\  \hline
    \end{tabular}

\vspace{-3mm}
\end{table}
\subsection{On the benefit of the proposed augmentation methods to label-efficient settings}

It is worth mentioning that our proposed augmentation techniques are orthogonal to which setting it is conducted. In this part, besides supervised 3D detection, we also investigate the effectiveness of our proposed augmentation in label-efficient settings that include semi-supervised and cross-domain scenarios.

\noindent{\textbf{Semi-supervised training.}} In semi-supervised learning, one of the common practices~\cite{deng2021unbiased, semi_consistency} is to regularize the output consistency of the unlabeled data under image manipulations. As for monocular 3D detection, we utilize our proposed augmentation to generate different views of unlabeled data and then feed them into mean-teacher architecture~\cite{deng2021unbiased, semi_consistency} to regularize the geometric consistency of their outputs. In terms of the different levels of manipulation, the regularization requires detectors to estimate consistent object dimension and yaw angle and predict depth that satisfied the geometric relationships.  

We conduct this case study on the KITTI dataset by using the ``Eigen-clean'' split~\cite{dd3d} with 14,940 images as the unlabeled subset and the training split as the labeled subset.
We provide the detailed setup of the mean-teacher framework on the supplementary material. Figure~\ref{fig:semi} shows the detection performance with different numbers of labeled data. Compared with the ``baseline'' that adopts the vanilla augmentation, the version with geometry-aware data augmentation obtains significant improvements when 500$\sim$1500 labeled data are sampled. Furthermore, when semi-supervised training is conducted with the unlabeled data, it achieves higher performance over the baseline version. This superior results demonstrate the potential of our augmentation techniques to reduce the labeling budget.

\noindent{\textbf{Cross-domain evaluation.}}
As stated in Section~\ref{sec:aug}, the geometric manipulations correspond to the shift of the camera configurations. We adopt a cross-domain evaluation to evaluate if the proposed augmentation techniques can enhance the detectors' robustness in real-scenario camera configuration shifts. Specifically, we conduct a KITTI to nuScenes evaluation, where the models are trained on the source domain (KITTI) and tested in the unseen target domain (nuScenes). On the KITTI and nuScenes datasets, the cameras' focal length and their receptive field are different. As shown in Table~\ref{tab:cross_domain}, the augmentation enhanced detector not only outperforms baseline in the in-domain scenario but also shows better robustness in the cross-domain situation. 



\section{Conclusion and Discussion}
In this work, we diagnosed the instability issues of monocular detectors under geometric shifts.
To alleviate the geometric inconsistency issues observed in the diagnosis, we proposed diverse augmentation techniques for regularizing the monocular object detectors.
Our work provides a new way to improve the 3D detection performance by generating more training data with preserving the geometric properties. With more diverse training data, the augmentation methods yield consistently improvement over state-of-the-art approaches on the KITTI and nuScenes datasets.

Except for the simple image perturbations, sophisticated augmentation techniques have already emerged in 2D object detection and 3D scene understanding for improving model robustness (\textit{e.g.,} mixup, novel view synthesis, sim-to-real, adversarial example, \textit{etc.}).
On the other hand, monocular 3D object detection also has its robustness issues (\textit{e.g.,} the perturbation of camera pitch and roll angle, occlusion, \textit{etc.}), which could be alleviated by customized data augmentation methods. 
We hope this paper will provide a baseline setup for future work in leveraging augmentation methods to enhance monocular 3D object detection. 
%
%

{\small
\bibliographystyle{ieee_fullname}
\bibliography{egbib}
}

\clearpage

\begin{center}
	\Large\textbf{Appendix}
\end{center}
\begin{figure*}[!htb]
    \centering
    \includegraphics[width=1.0\textwidth]{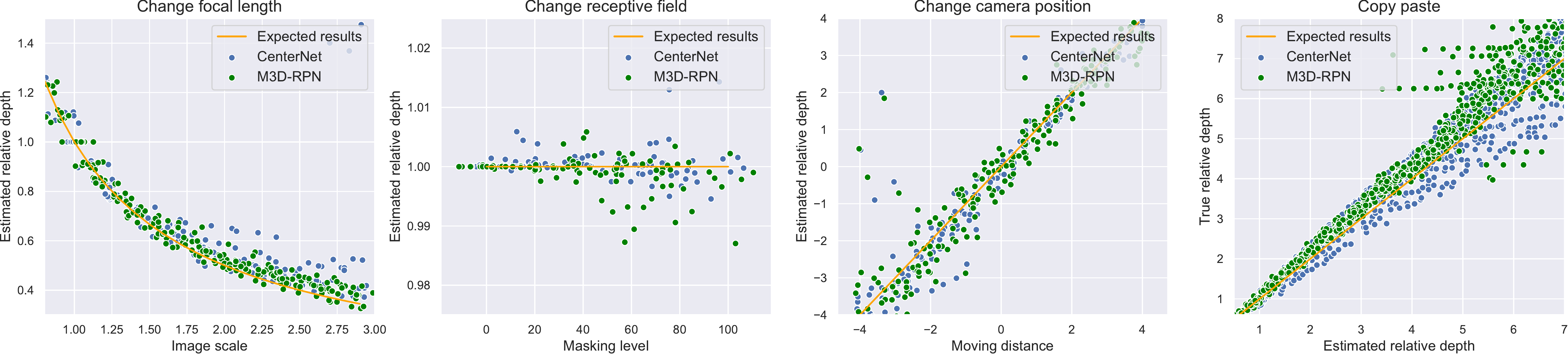}
    \caption{Empirical analysis of augmentation enhanced detectors under geometric manipulations. }
    \label{fig:aug_analysis}
\end{figure*}

The content of supplementary material is organized as follows:
\begin{itemize}
    \item Section~\ref{sec:exp} conducts more evaluations of the geometry-aware strategy. 
    \item Section~\ref{sec:exp_setup} and~\ref{sec:copy_paste_detail} introduce the implementation details of the data augmentation methods. 
    \item Section~\ref{sec:semi_supervised} presents the details of semi-supervised training settings. 
\end{itemize}

\section{More experimental results}
\label{sec:exp}
We display the experimental results of our geometry-aware augmentation in Table~\ref{tab:eff_geo}. As illustrated, our augmentation methods effectively enhance the model robustness under different kinds of perturbation. Compared to the vanilla version in the main paper, the performance of augmentation enhanced detectors is much better in the perturbation settings. 
\subsection{Stability of augmentation enhanced detectors}
\label{sec:exp_stability_aug}
\begin{table}[!htb]
    \centering
    \caption{Experimental results of Anchor-based (M3D-RPN) and  Anchor-free (CenterNet) detectors under different manipulation techniques. Except the baseline setting, we replace the ground-truth with estimated results. For example, ``Depth*`` denotes replacing the ground truth depth with the estimation and setting all other components with ground truth. (Results of $AP|_{40}$ with IoU$\geq$0.5 on car (easy) are reported.)}
    \tabcolsep2pt
    \begin{tabular}{ll|ccccc} \hline
    Network&    Method & Base & Depth* & Dim* & Pos*  \\ \hline
    \multicolumn{1}{c}{\multirow{5}{*}{M3D-RPN}} &  
    Origin &  65.9 & 70.2 & 99.2 & 99.0 &\\ 
        &Random scale &60.1 & 68.1 & 98.5 & 98.6    \\ 
        &Random crop & 59.2 & 62.3 & 96.6 & 96.7\\
        &Moving cam & 52.8 & 62.8 & 93.9 & 92.1\\
        & Copy-paste & 53.2 & 58.3 & 89.4 & 98.2\\ \hline
    \multicolumn{1}{c}{\multirow{5}{*}{CenterNet}} & 
        Origin  & 60.3 & 65.3 & 99.1 & 99.0 \\
        & Random scale & 55.3 & 62.3 & 98.9 & 98.8  \\ 
        & Random crop& 58.8 & 64.2 & 97.3 & 98.2 \\
        & Moving cam & 50.3 & 59.8 & 91.7 & 88.6 \\
        & Copy-paste & 49.2 & 52.1  & 90.0 & 98.8 \\ \hline
    \end{tabular}
    \vspace{-5mm}
    \label{tab:eff_geo}
\end{table}
In Figure~\ref{fig:aug_analysis}, we also display the empirical analysis we conducted in Section [4] to evaluate whether our proposed data augmentation methods can enhance the stability. Compared with the baseline results in Figure[4], the results from the augmentation enhanced detectors are more fixed with the expected results and have less deviation.

\section{nuScenes datasets}
\begin{table}[!htb]
    \vspace{-3mm}

    \centering
    \caption{Experimental results of the anchor-free detector on the nuScenes validation set. }
\vspace{-3mm}
    \tabcolsep2pt

    \begin{tabular}{ll|cccc} \hline
   Pretrained  & Setting & mAP$\uparrow$&mATE$\downarrow$&mASE$\downarrow$& NDS$\uparrow$ \\ \hline
   \multirow{2}{*}{ImageNet}  &   Vanilla aug & 33.2 &  0.69 & 0.28  & 38.4      \\  
      &  Geo aug & 34.5   & 0.68 & 0.27   & 39.4      \\ \hline
     \multirow{2}{*}{\makecell{ImageNet+\\DDAD}} & Vanilla aug & 34.6 & 0.67 & 0.27 &  39.4 \\ 
     & Geo aug &  35.6 & 0.66 & 0.26 & 40.6\\ \hline
    \end{tabular}
    \label{tab:nuScenes}
        \vspace{-3mm}
\end{table}

We first introduce the detailed experimental setting on the nuScenes dataset and provide additional results of CenterNet~\cite{zhou2019objects} with different augmentation strategies. 
In the nuScenes dataset, we utilize the AdamW optimizer to train the models with 48 epochs. The initial learning rate is 4e-2 and downscaled with 0.1 in the 32$^{th}$ and 44$^{th}$ epoch. To save the memory occupation, we rescale the input resolution from $1600\times 900$ to $1200 \times 675$ in both training and inference, where the batch size is set as 80 during training. 

In Table~\ref{tab:nuScenes}, we provide the experimental results of CenterNet with different pre-trained weights. DDAD~\cite{dd3d} denotes the private datasets reported in DD3D~\cite{dd3d}. We utilize the provide pre-trained models to initialize the modified DLA-34 backbone in the detection model. Experimental results illustrate the effectiveness of our geometry-aware strategy in a stronger baseline setting.

%


\section{Details about geometry-aware data Hyper-parameters in data augmentation}
\label{sec:exp_setup}
The hyper-parameters for the data augmentation are represented as follows:
1). Random Crop: we randomly crop the image with size of 960$\times$320. 2). Random Scale: we randomly resize the image with a range from 0.8 to 1.2, with fixing the size ratio. 3). Camera position: To alleviate generate artifact, we control the change distance of camera position from -5 to 5 meters. 4). Copy-paste: We first utilizes an instance segmentation method~\cite{wu2019detectron2} to crop the foreground objects with around 12,581 instances. After that, we randomly select two cropped instances and insert them into every training image with sampling new depth from 0 - 70. 

\section{Details of Copy-paste augmentation method}
\label{sec:copy_paste_detail}
\noindent{\textbf{Generating bounding boxes}}
For the step 7 in the Algorithm 1, we utilize the acquired object dimension, location, orientation to get the final bounding boxes 2D coordinates. The procedure is similar in ~\cite{liu2019deep}. We first calculate the rotation matrix R with using the egocentric orientation angle:
\begin{align}R = \left(
    \begin{array}{ccc}
        \cos \theta & 0 & \sin \theta  \\
        0  &  1 & 0 \\
        -\sin \theta & 0 & \cos \theta 
    \end{array} \right).
\end{align}
The 8 corner points in the object coordinate is:
\begin{align}
    P^{3d}_{4\times 8} = \left(\begin{smallmatrix}
        \frac{L}{2} &\frac{L}{2} & -\frac{L}{2} & -\frac{L}{2} & \frac{L}{2} & \frac{L}{2} & -\frac{L}{2} & -\frac{L}{2}   \\
        0 & 0 &0 & 0 & -H & -H & -H & -H  \\
        \frac{W}{2} &  -\frac{W}{2}  & -\frac{W}{2}  & \frac{W}{2}  & \frac{W}{2}  & -\frac{W}{2}  & -\frac{W}{2}  & \frac{W}{2}  \\
        1 & 1 & 1 & 1 & 1 & 1 & 1 & 1 
    \end{smallmatrix}\right). \nonumber
\end{align}
For the coordinate of point i, it is calculated as follows:
\begin{align}
    P^{2d}_{3\times 8}= K_{3\times4}\left(\begin{array}{cc}
        R & T  \\
        0^T & 1 
    \end{array}\right)P^{3d}_{4\times 8},
\end{align}
where T is the 3D location matrix with $\left[X, Y, Z\right]$, and $P^{2d}_{3 \times 8}$ is the coordinates in the images. 

\section{Implementation details in the semi-supervised setting}
\label{sec:semi_supervised}
In this paper, we adopt the mean-teacher framework~\cite{semi_consistency} to regularize the output consistency of monocular detectors. Following existing work~\cite{semi_consistency}, we first select the candidate bounding boxes based on the pooling module (in CenterNet) or nms module (in M3D-RPN). Then we select the candidates with confidence score larger than 0.7 for regularization. The teacher network is the momentum version of the student network with factor of 0.9. We fed the teacher network with origin image and the student network with augmented images. The weight of the regularization loss is set as 1. 



 \end{document}